\documentclass[conference]{IEEEtran}
\IEEEoverridecommandlockouts

\usepackage[utf8]{inputenc}
\usepackage[pdftex]{graphicx}

\usepackage{textcomp}

\usepackage{amsmath,amssymb,amsfonts,url,multirow}
\usepackage{algorithmicx}
\usepackage{algpseudocode}
\usepackage{cases}
\usepackage{ulem}
\usepackage{epstopdf}
\usepackage{listings}
\usepackage{array}
\usepackage{makecell}
\usepackage{tabularx}
\usepackage{multicol}

\usepackage[x11names,table]{xcolor}
\usepackage{colortbl}

\usepackage{caption}
\usepackage{booktabs}

\include{timeline}

\usepackage[edges]{forest}

\usepackage{pifont}

\usepackage{hyperref}

\def\BibTeX{{\rm B\kern-.05em{\sc i\kern-.025em b}\kern-.08em
    T\kern-.1667em\lower.7ex\hbox{E}\kern-.125emX}}

\newcommand\PRR{\textrm{PRR}}

\newcommand\Wprr{\textrm{W}_\textrm{PRR}}
\newcommand\Whistory{\textrm{W}_\textrm{history}}

\ifCLASSOPTIONcompsoc
	\usepackage[nocompress]{cite}
	\usepackage[caption=false,font=normalsize,labelfont=sf,textfont=sf]{subfig}
\else
	\usepackage{cite}
	\usepackage[caption=false,font=footnotesize]{subfig}
\fi

\definecolor{awesome}{rgb}{1.0, 0.6, 0.0}

\usepackage[nopostdot,nomain,nonumberlist,acronym]{glossaries}
\setglossarystyle{super}

{\centering\tablehead{}\tabletail{}%
	\begin{supertabular}{lp{\glsdescwidth}}}%
	{\end{supertabular}}%

\makeglossaries

\newacronym{LQE}{LQE}{Link Quality Estimation}
\newacronym{PRR}{PRR}{Packet Reception Ratio}
\newacronym{WSN}{WSN}{Wireless Sensor Network}
\newacronym{SNR}{SNR}{Signal-to-Noise Ratio}
\newacronym{BER}{BER}{Bit Error Rate}
\newacronym{LQI}{LQI}{Link Quality Indicator}
\newacronym{RSSI}{RSSI}{Received Signal Strength Indicator}
\newacronym{WMEWMA}{WMEWMA}{Window Mean with an Exponentially Weighted Moving Average}
\newacronym{KDP}{KDP}{Knowledge Discovery Process}
\newacronym{CDF}{CDF}{Cumulative Distribution Function}
\newacronym{4B}{4B}{Four-Bit}
\newacronym{4C}{4C}{Foresee}
\newacronym{PSR}{PSR}{Packet Success Ratio}
\newacronym{RSS}{RSS}{Received Signal Strength}
\newacronym{ETX}{ETX}{Expected Transmission count}
\newacronym{RNP}{RNP}{Required Number of Packets}
\newacronym{SGD}{SGD}{Stochastic Gradient Descent}
\newacronym{FLI}{FLI}{Fuzzy-logic Link Indicator}
\newacronym{F-LQE}{F-LQE}{Fuzzy-logic based LQE}
\newacronym{WNN-LQE}{WNN-LQE}{Wavelet Neural Network based LQE}
\newacronym{MSE}{MSE}{Mean Squared Error}
\newacronym{RUS}{RUS}{Random Under-Sample}
\newacronym{ROS}{ROS}{Random Over-Sample}
\newacronym{TCP}{TCP}{Transmission Control Protocol}
\newacronym{KDD}{KDD}{Knowledge Discovery and Data mining}
\newacronym{ML}{ML}{Machine Learning}
\newacronym{AI}{AI}{Artificial Intelligence}

\newacronym{SVM}{SVM}{Support Vector Machine}
\newacronym{SVR}{SVR}{Support Vector Regressor}
\newacronym{LQ}{LQ}{Link Quality}
\newacronym{NLQ}{NLQ}{Neighbor Link Quality}
\newacronym{MAE}{MAE}{Mean Absolute Error}
\newacronym{RMSE}{RMSE}{Root-Mean-Square Error}
\newacronym{PER}{PER}{Packet Error Rate}
\newacronym{ROC}{ROC}{Receiver Operating Characteristic}

\begin{document}

\title{Learning to Fairly Classify the Quality of Wireless Links}

\author{Gregor Cerar$^{\ast\dagger}$,
     	Halil Yetgin$^{\ast\ddagger}$,
        Mihael Mohor\v ci\v c$^{\ast\dagger}$,
        Carolina Fortuna$^{\ast}$\\
$^{\ast}$Department of Communication Systems, Jo{\v z}ef Stefan Institute, SI-1000 Ljubljana, Slovenia.\\
$^{\dagger}$Jo{\v z}ef Stefan International Postgraduate School, Jamova 39, SI-1000 Ljubljana, Slovenia.\\
$^{\ddagger}$Department of Electrical and Electronics Engineering, Bitlis Eren University, 13000 Bitlis, Turkey.\\
\{gregor.cerar $\mid$ halil.yetgin $\mid$ miha.mohorcic $\mid$ carolina.fortuna\}@ijs.si    
}

\maketitle

\begin{abstract}
	Machine learning (ML) has been used to develop increasingly accurate link quality estimators for wireless networks. However, more in depth questions regarding the most suitable class of models, most suitable metrics and model performance on imbalanced datasets remain open. In this paper, we propose a new tree based link quality classifier that meets high performance and fairly classifies the minority class and, at the same time, incurs low training cost. We compare the tree based model, to a multilayer perceptron (MLP) non-linear model and two linear models, namely logistic regression (LR) and SVM, on a selected imbalanced dataset and evaluate their results using five different performance metrics. Our study shows that 1) non-linear models perform slightly better than linear models in general, 2) the proposed non linear tree-based model yields the best performance trade-off considering F1, training time and fairness, 3) single metric aggregated evaluations based only on accuracy can hide poor, unfair performance especially on minority classes, and 4) it is possible to improve the performance on minority classes, by over $40\%$ through feature selection and by over $20\%$ through resampling, therefore leading to fairer classification results.
\end{abstract}

\IEEEpeerreviewmaketitle


\section{Introduction}
\label{sec:intro}
Machine learning (ML) is becoming an increasingly popular way of solving various problems in communications in general and wireless networks in particular. Data driven link quality estimation (LQE) techniques where the researchers manually developed models have been proposed over the last two decades~\cite{nguyen1996trace,balakrishnan1998explicit,woo2003taming}. More recently, the manual model development is being automated, by using machine learning algorithms that approximate the distribution of the underlying random variable and are thus able to learn the quality of a link~\cite{sun2017wnn,demetri2019automated}. 

LQE models developed using ML algorithms can estimate the quality of a link in a continuous-valued space by means of performing regression~\cite{liu2011foresee, sun2017wnn, ancillotti2017reinforcement, okamoto2017machine, bote2018online}. Alternatively, if they estimate the link quality in a discrete-valued space, ML performs classification~\cite{liu2014temporal, rekik2015fli,luo2019link, demetri2019automated}. By analyzing the existing body of literature developing classification models for LQE, we notice two types of approaches; i) \textit{binary- or two-class}, ii) \textit{multi-class}. 

The \textit{binary- or two-class} approach, can be found in~\cite{guo2013fuzzy, liu2014temporal, rekik2015fli} while \textit{multi-class} approach appears in~\cite{rehan2016machine, shu2017research, audeoud2018quick, luo2019link, demetri2019automated}, where \cite{rehan2016machine,audeoud2018quick} use a three-class, \cite{boano2010triangle} utilizes a four-class, \cite{shu2017research, luo2019link} rely on a five-class, and \cite{demetri2019automated} leverages a seven-class output. These applications are leveraged for the categorization and estimation of the future link state, which is expressed through labels/classes and it is not always clear from the related work how the authors select the number of classes. The binary-class works seem to be motivated by the application requirements, particularly of a multi-hop routing protocol that needs to know whether a link is reliable or not. The three-class approach seems to be motivated by the non-linear S-shaped curve with three regions specified for wireless links~\cite{baccour2012radio}. The seven-class output is motivated by the geographical environment over which the wireless network operates considering the application of coverage estimation~\cite{demetri2019automated}. 

An important aspect that is not previously considered in LQE classification and possibly neither in general classification problems for wireless communications is the fairness of the ML models developed for classification. However, maintaining fairness in multi-class classification problems has been a challenging issue, especially when an imbalanced dataset is considered~\cite{Zhang9117188}. To exemplify the significance of classification unfairness in real-life scenarios, Chouldechova~\textit{et al.}~\cite{Chouldechova2017} show evidence of racial bias in the recidivism prediction tool, in which white defendants are less likely to be classified as high-risk than black defendants and Obermeyer~\textit{et al.}~\cite{Obermeyer447} show biases in the health care decision-making system in which black patients who are captured by the algorithm at the same risk level are sicker than white patients. Resembling these real-life classification problems to the wireless communication links, when no good links are available and the classifier is unable to recognize intermediate links as these usually belong to the minority class that is unfairly discriminated, the communication might be hindered by selecting a bad link. Therefore, it is important to justify whether the decision made by a ML model is fair to all considered link quality classes. Against this background, we propose a decision tree-based ML model for LQE with the goal of attaining fairness between link quality classes, albeit with the least possible accuracy compromise, and compare this accuracy/fairness performance trade-off to other existing ML models.

From the analysis of the literature discussed above, we draw the following observations:
\paragraph*{Observation-1} ML based classification studies that use linear ML methods, such as logistic regression alongside non-linear methods, such as neural networks~\cite{liu2011foresee,liu2014temporal} reveal small performance differences in the range of few percentage points on the three zone S-like shaped link quality curve. According to~\cite{sun2017wnn}, link quality tends to be a non-linear function, thus non-linear models are likely to perform better for LQE. However, this aspect is not systematically investigated in the literature. 

\paragraph*{Observation-2} Most of the related works on classification evaluate their performance using the $accuracy$ metric and perhaps some other application-specific metric, such as routing tree stability or depth. Notable exceptions are~\cite{luo2019link, demetri2019automated}, where the authors present a full confusion matrix to be able to assess which classes are well discriminated by the model and which are often confused. However, it is well-known in the ML communities that accuracy is a misleading metric, especially for imbalanced datasets~\cite{jeni2013facing}, where it can hide bias or unfairness towards the minority class~\cite{Zhang9117188}.

\paragraph*{Observation-3} The authors of~\cite{luo2019link} provide a great level of details in their methodology and in their results. Their confusion matrices reveal very strong performance on certain classes and higher confusion on others. Relatively poorer performance on intermediate classes may be due to the class imbalance on the training data. However, we are unable to see if this is the case with their training data and by looking at their process, no countermeasures, e.g. resampling techniques seem to be adopted as a remedy.

Following the three listed observations, we identify opportunities to contribute and extend the existing body of work on LQE using ML based classification, as follows.

\begin{itemize}
	\item We propose a new tree based link quality classifier that meets high classification performance and fairly classifies also the minority class while, at the same time, incurring low training cost. 
	
	\item We compare the proposed tree based model, to a multilayer perceptron (MLP) non-linear model and two linear models, namely LR and SVM, on a selected imbalanced dataset and show that the proposed model takes about 90 times less training time compared to MLP and the performance compromise is less than $\approx1\%$.
	
	\item We adopt standard metrics from the ML community to evaluate the performance of our classifier. In addition to $accuracy$, we also use $precision$, $recall$, $F1$ and, where necessary, the detailed $confusion$ $matrix$ based on which all the other metrics are computed. To date, no other LQE classification work considered all five different metrics for a thorough performance evaluation that also considers per class fairness.
	
	\item We explicitly study and evaluate ways to improve minority class discrimination on imbalanced datasets for the sake of a fair classification performance on all link quality classes. For this purpose, we select a publicly available wireless dataset that is suitable for developing an LQE classifier and is imbalanced. 
\end{itemize}

The rest of this paper is structured as follows. Section~\ref{sec:rw} summarizes related work while Section~\ref{sec:learning} defines the learning problem, including a preliminary for linear and non-linear ML-based models, dataset selection and methodology. Section~\ref{sub:feature-selection} elaborates on selecting the best features for training a model with high performance and fair per-class discrimination capabilities. Section~\ref{sub:resampling} studies how to compensate for the class imbalance in the dataset to further improve per class fairness while Section~\ref{sub:building-model} evaluates the performance of the proposed model. Finally, in Section~\ref{sec:conclusion} summarizes the paper and identifies future directions.

\section{Related work}\label{sec:rw}
To the extent of our knowledge, this is the first attempt to develop a ML-based LQE model that considers classification fairness among the accounted wireless link quality classes. Moreover, there is only a paucity of contributions considering decision tree-based ML algorithms for LQE.

One of the first ML models for LQE is proposed by Liu~\textit{et al.}~\cite{liu2011foresee}, in which they use the 4C algorithm to train three ML models based on na\" ive Bayes, neural networks, and logistic regression algorithms, which ultimately produces a multi-class output. Subsequently, Liu~\textit{et al.}~\cite{liu2014temporal} extend their work to an online ML model, namely TALENT, where the model built on each device adapts to newly generated data points instead of being pre-computed on a server, and consequently yields a binary threshold-based output. 

Similarly, Shu~\textit{et al.}~\cite{shu2017research} use the support vector machine (SVM) algorithm to develop a five-class link quality model, while Okamoto~\textit{et al.}~\cite{okamoto2017machine} use an online learning algorithm called adaptive regularisation of weight vectors for learning to estimate throughput from images, and then Bote-Lorenzo ~\textit{et al.}~\cite{bote2018online} train online perceptrons, online regression trees, fast incremental model trees, and adaptive model rules. The latter two models consider continues-valued output, which means that they are simply constrained by numerical precision due to regression. Demetri~\textit{et al.}~\cite{demetri2019automated} propose a seven-class SVM classifier to estimate LoRa network coverage, using multiple input metrics to train the classifier, including multispectral aerial imagery. Surprisingly, the only reinforcement learning-based approach for LQE is found in~\cite{ancillotti2017reinforcement}, where the authors train a greedy algorithm with multiple input metrics to estimate PRR as a continuous-valued output in terms of protocol improvement in mobility scenarios. 

Furthermore, two LQE models using deep learning algorithms have been proposed, where the first model~\cite{sun2017wnn} introduces a new LQE metric for estimating link quality in smart grid environments that relies on SNR while producing a continuous-valued PRR output. In the other model, Luo~\textit{et al.}~\cite{luo2019link} incorporate multiple input metrics and train neural networks to discriminate an LQE model with five classes. 

None of the aforementioned works dealing with multi-class classification problems consider fairness among accounted classes and decision tree-based ML algorithms. Only in our recent work~\cite{cerar2020}, we evaluate the performance of logistic regression, three-based, ensemble, and multilayer perceptron algorithms for LQE with a three-class output and show that feature engineering has a larger impact on the final LQE model performance than the choice of ML algorithms. However, the fairness among the considered classes was not analysed in this particular work.

\section{Definition of the learning problem}
\label{sec:learning}
We aim to learn to discriminate among the widely-used three-class distinction model~\cite{baccour2012radio}, i.e., \textit{good}, \textit{intermediate} and \textit{bad} classes for a link. To achieve this, we leverage the selected dataset and the identified linear and non-linear ML algorithms, and train the algorithms with a subset of the available data. This way, a model that is able to discriminate among the three target classes is developed and its performance is then evaluated on the remaining data. To conduct our study and evaluate the performance of the proposed DTree and the other three models, we use the standard approach for developing a classifier: we first perform data pre-processing, then continue with model training and selection. 

\subsection{Linear and non-linear ML-based models}
\label{sec:linearnonlinear}
Machine learning algorithms are suitable for automatically approximating the underlying distribution that generated a set of measurements. They are particularly useful when there is no analytical formula that models the phenomenon generating the distribution and a large number of empirical observations can be collected. If the measurements are closer to a non-linear function, than non-linear ML algorithms such as decision trees are more suitable for approximating them. Otherwise, linear models such as logistic regression are preferred due to their simplicity and relatively lower computational complexity~\cite{Bishop2006}. 

For linear ML-based LQE model development, we consider \textit{logistic regression} as a subset of the general linear regression and \textit{support vector machine (SVM) with linear kernel}. A logistic regression function enforces the output of the linear function to lie between the value of 0 and 1, where the classification (labeling) of link quality is conducted based on a predetermined threshold. This can be achieved by maximizing the probability of a random data point to be correctly classified relying on maximum likelihood, gradient descent or other optimization algorithms. Similarly, SVM with linear kernel produces a hyperplane or a line (depending on the number of features) that precisely classifies data points. The main idea of the SVM is to maximize the margin between respective data points that are closer to the hyperplane~\cite{Bishop2006}.

On the other hand, the considered non-linear ML-based LQE models are developed using \textit{decision trees} (DTree) and \textit{multilayer perceptron (MLP)}. A decision tree represents a non-linear mapping of the independent and dependent variables, which can be utilized for classifying data that is difficult to separate with linear methods~\cite{Bishop2006}. MLP represent a subset of feedforward artificial neural networks composed of at least three layers of nodes, each of which is a neuron that utilizes a non-linear activation function. MLP can classify data that is not linearly distinguishable~\cite{Bishop2006}.

\begin{figure}[!htb]
	\centering
	\includegraphics[width=0.85\linewidth]{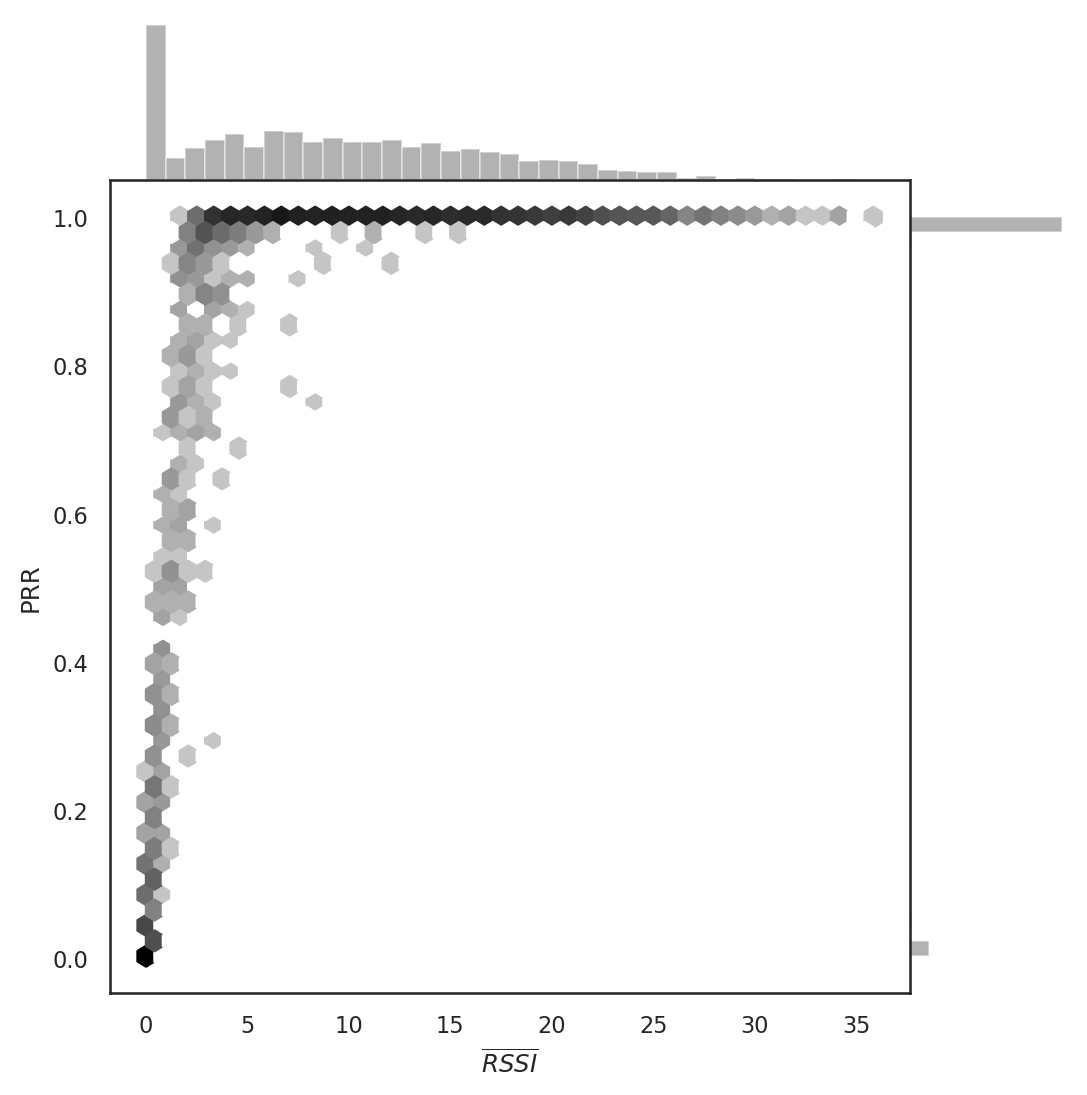}
	\caption{Link between Packet Reception Ratio (PRR) and average Received Signal Strength Indicator ($\overline{\textrm{RSSI}}$) on logarithmic scale for Rutgers trace-set.}
	\label{fig:rutgers-prr-distribution}
\end{figure}

\subsection{Trace-set selection}
\label{sec:rutgers}
As discussed in Section \ref{sec:intro}, the third aspect of our investigation requires an imbalanced dataset that is suitable for training a ML based LQE. We also prefer a publicly available dataset so that the research can be easily replicated. We have identified a number of such publicly available datasets, namely Roofnet~\cite{aguayo2004link}, Rutgers~\cite{kaul2006creating}, ``packet-metadata''~\cite{fu2015experimental}, University of Michigan~\cite{chen2011robust}, EVARILOS~\cite{van2015platform} and Colorado~\cite{anderson2009impact}.

Roofnet~\cite{aguayo2004link} is a well known WiFi-based trace-set and contains the largest number of data points among the identified trace-sets, however packet reception ratio~(PRR), as a target training metric for the classifier, can only be computed as an aggregate value per link without the knowledge of how the link quality varied over time. Rutgers is smaller than Roofnet, however is large enough to train a ML model and is appropriately formed for our purpose. The trace-set for each node contains raw received signal strength indicator~(RSSI) value along with the sequence number. 

Upon closer investigation for the remaining trace-sets, we concluded that they are not suitable for our intended purpose. The ``packet-metadata''~\cite{fu2015experimental} comes with a plethora of features convenient for LQE research. In addition to the typical LQI and RSSI, it provides information about the noise floor, transmission power, dissipated energy as well as several network stacks and buffer related parameters. However, packet loss can only be observed in rare cases with very small packet queue length.

The trace-set from the University of Michigan~\cite{chen2011robust} is somewhat incomplete and suffers from an inconsistent data format containing lack of units, missing sequence numbers and inadequate documentation. The two EVARILOS trace-sets~\cite{van2015platform} are mainly well-formatted, whereas each contains fewer than 2,000 entries. In the Colorado trace-setColorado~\cite{anderson2009impact}, the diversity of the link performance is missing as all links seem to exhibit less than 1\% packet loss. 

After careful consideration we selected the ``Rutgers trace-set''~\cite{kaul2006creating} as the candidate dataset for this work. The dataset was created using the ORBIT testbed and includes 4,060 distinct link traces, which are gleaned from 812 unique links with 5 different noise levels, i.e., 0, -5, -10, -15 and -20 dBm. Readily available trace-set features include raw RSSI, sequence numbers, source node ID, destination node ID and artificial noise levels. The packets are sent every 100 milliseconds for a period of 30 seconds, therefore, each trace is composed of 300 packets. Besides, based on the specifications of the radio used, each RSSI value is defined between 0 and 128, where the value of 128 indicates an error and is therefore invalid. A statistical analysis of the Rutgers trace-set reveals that 960 link traces out of 4,060 (23.65\%) are entirely empty indicating no packets were received, and that a total of 1,218,000 packets were sent and only 773,568 (63.51\%) were correctly received. 

We plot in Fig.~\ref{fig:rutgers-prr-distribution} the relationship between RSSI and the PRR computed based on the available sequence numbers. The darker hexagonal areas of Fig.~\ref{fig:rutgers-prr-distribution} indicate that the majority of links are of either ``poor quality'' (bottom-left) or ``good quality'' (top), while gray areas are of ``intermediate quality''. The bars on the right hand side of the figure show the imbalanced nature of the dataset, more precisely, 61\% of the links are \textit{good}, 34\% are \textit{bad} and only 5\% \textit{intermediate}.

\begin{table}
	\caption{Global parameters for ML-based LQE models.}
	\label{tab:parameters}
	\centering
	\begin{tabular}{@{}ll@{}}
		\toprule
		\bfseries Step/Parameter
		& \bfseries Default value
		\\
		
		\midrule
		
		Missing data
		& Domain knowledge (zero-fill)
		\\
		
		History window size (W$_\text{history}$)
		& 10
		\\
		
		Prediction window size (W$_\text{PRR}$)
		& 10
		\\
		
		Features set
		& RSSI, $\overline{\text{RSSI}}_{10}$, RSSI$_\text{SD,10}$
		\\
		
		Resampling strategy
		& Random oversampling (ROS)
		\\
		
		Link quality labels
		& \textit{Good}, \textit{intermediate}, \textit{bad}
		\\
		
		\midrule
		
		\multirow{5}{*}{Globally used ML algorithms}
		& \textbf{Linear:}  Linear (Logistic) \\
		& \textbf{Non-linear:} Decision trees (DTree) \\
		& \quad with tree depth limited to 4, \\
		& \quad the min. samples per node set to 50 \\ 
		
		\midrule
		
		Cross-validation strategy
		& Randomize \& 10-times Stratified K-Fold
		\\
		
		\bottomrule
	\end{tabular}
\end{table}

\subsection{Experimental details}
As a baseline reference model, we select the \textit{majority classifier}, which in our case, classifies all the links in good quality class. In order to evaluate the most suitable ML-based LQE model, we utilize accuracy, precision, recall and F1 metrics. For our analysis, we include per class score values in parentheses for precision, recall and F1 values as in the following order: \textit{good}, \textit{intermediate} and \textit{bad}. Then, these values in parenthesis are averaged using a \textit{weighted average value per class} method to obtain precision, recall and F1 values, respectively. For the sake of providing a fair comparison, before any ML-based LQE model is developed, the dataset is shuffled and 10-times stratified K-Fold is employed to produce estimated classes~\cite{arlot2010survey}. For the development of ML-based LQE models, we utilize the global parameters of Table~\ref{tab:parameters} throughout the paper, unless stated otherwise. $\Whistory$ in Table~\ref{tab:parameters} represents the historical window that is utilized for calculating the features and $\Wprr$ depicts the prediction window that is used for identifying the link quality labels. $\overline{\text{RSSI}}_{10}$ represents the averaged RSSI over 10 packets and $RSSI_\text{SD,10}$ represents the standard deviation of the RSSI over 10 packets. Missing values in the Rutgers are filled using the zero-filling technique, as outlined in Table~\ref{tab:parameters}.

\begin{table*}
	\caption{Comparison of various sets of features using linear and non-linear ML algorithms.}
	\label{tab:features}
	\renewcommand{\arraystretch}{1.34}
	\centering
	\begin{tabular}{l l l l l l}
		\toprule
		\bfseries Algorithm
		& \bfseries Feature set
		& \bfseries Acc. [\%]
		& \bfseries Precision [\%] 
		& \bfseries Recall [\%]
		& \bfseries F1 [\%]
		\\\midrule

		\multirow{6}{*}{Linear (Logistic)}
		& RSSI
		& 74.4
		& 77.3 (86.3, 81.4, 64.3)
		& 74.4 (92.8, 30.9, 99.3)
		& 70.8 (89.5, 44.8, 78.1)
		\\

		& $\overline{\text{RSSI}_{10}}$
		& 89.7
		& 89.8 (92.6, 90.0, 86.9)
		& 89.7 (93.8, 77.8, 97.5)
		& 89.5 (93.2, 83.5, 91.9)
		\\

		& RSSI$_{\text{SD},10}$
		& 77.1
		& 78.4 (82.8, 64.3, 88.1)
		& 77.1 (55.6, 79.3, 96.6)
		& 76.6 (66.5, 71.0, 92.1)
		\\

		& RSSI, $\overline{\text{RSSI}}_{10}$, RSSI$_{\text{SD},10}$
		& \textbf{92.2}
		& \textbf{92.3} (97.1, 90.2, 89.6)
		& \textbf{92.2} (93.9, 86.0, 96.7)
		& \textbf{92.2} (95.5, 88.0, 93.0)
		\\
		
		& $\Delta$RSSI (``left'' derivative)
		& 43.7
		& 31.3 (52.4, 0.0, 41.5)
		& 43.7 (31.6, 0.0, 99.4)
		& 32.7 (39.4, 0.0, 58.5)
		\\
		
		& $\overline{\textrm{RSSI}}_{10}^{\{-4, -3, -2, -1, 1, 2, 3, 4\}}$
		& 80.0
		& 80.0 (93.5, 72.0, 74.4)
		& 80.0 (92.3, 65.4, 82.3)
		& 79.9 (92.9, 68.6, 78.1)
		\\
		
		\midrule

		\multirow{6}{*}{Non-linear (DTree)}
		& RSSI
		& 75.1
		& 77.5 (92.2, 75.8, 64.3)
		& 75.1 (87.8, 38.2, 99.3)
		& 72.9 (90.0, 50.8, 78.1)
		\\

		& $\overline{\text{RSSI}}_{10}$
		& 91.6
		& 91.6 (94.5, 87.4, 93.1)
		& 91.6 (91.7, 87.5, 87.4)
		& 91.6 (93.1, 57.4, 94.3)
		\\

		& RSSI$_{\text{SD},10}$
		& 80.8
		& 80.7 (78.3, 71.3, 92.6)
		& 80.8 (72.7, 74.1, 95.6)
		& 80.7 (75.4, 72.7, 94.1)
		\\

		& RSSI, $\overline{\text{RSSI}}_{10}$, RSSI$_{\text{SD},10}$
		& \textbf{93.2}
		& \textbf{93.2} (96.2, 90.4, 93.0)
		& \textbf{93.2} (94.8, 89.0, 95.6)
		& \textbf{93.2} (95.5, 89.7, 94.3)
		\\

		& $\Delta$RSSI (``left'' derivative)
		& 60.3
		& 63.5 (69.6, 65.7, 55.2)
		& 60.3 (44.7, 37.4, 98.8)
		& 57.6 (54.4, 47.7, 70.8)
		\\

		& $\overline{\textrm{RSSI}}_{10}^{\{-4, -3, -2, -1, 1, 2, 3, 4\}}$
		& 80.0
		& 79.9 (93.0, 72.3, 74.4)
		& 80.0 (92.8, 64.8, 82.3)
		& 79.8 (92.9, 68.4, 78.1)
		\\

		\bottomrule
	\end{tabular}
\end{table*}

\section{The influence of feature selection on performance and fairness}
\label{sub:feature-selection}
Feature selection is the step in data preprocessing concerned with determining unprocessed features or creating synthetic features for the training of ML algorithms. Features can be conducted manually or produced by the aid of algorithms. The training feature available in our dataset is the raw Received Signal Strength Indicator (RSSI) value and the other is the sequence number that can be exploited for the limited time series analysis, and computation of PRR, on which the link quality classes depend. The arbitrary values associated to distinct classes, which were also set in~\cite{baccour2012radio}, are defined in the form of the following rule:

\begin{equation}\label{eq:prr-to-class}\small
y = f(\PRR) = \begin{cases}
\text{bad}, & \text{if } \PRR \leq 0.1 \\
\text{intermediate}, & \text{otherwise} \\
\text{good}, & \text{if } \PRR \geq 0.9,
\end{cases}
\end{equation}

\begin{equation}\label{eq:target-class}\small
\mathbf{y} = [y_1, y_2, \dots, y_n],\quad\forall{y}\in\{\text{bad},\hspace{0.2mm} \text{intermediate},\hspace{0.2mm} \text{good}\}.
\end{equation}

\begin{figure*}[!t]
	\centering		
	\subfloat[Per class precision values for logistic regression.]{\includegraphics[width=0.4\linewidth]{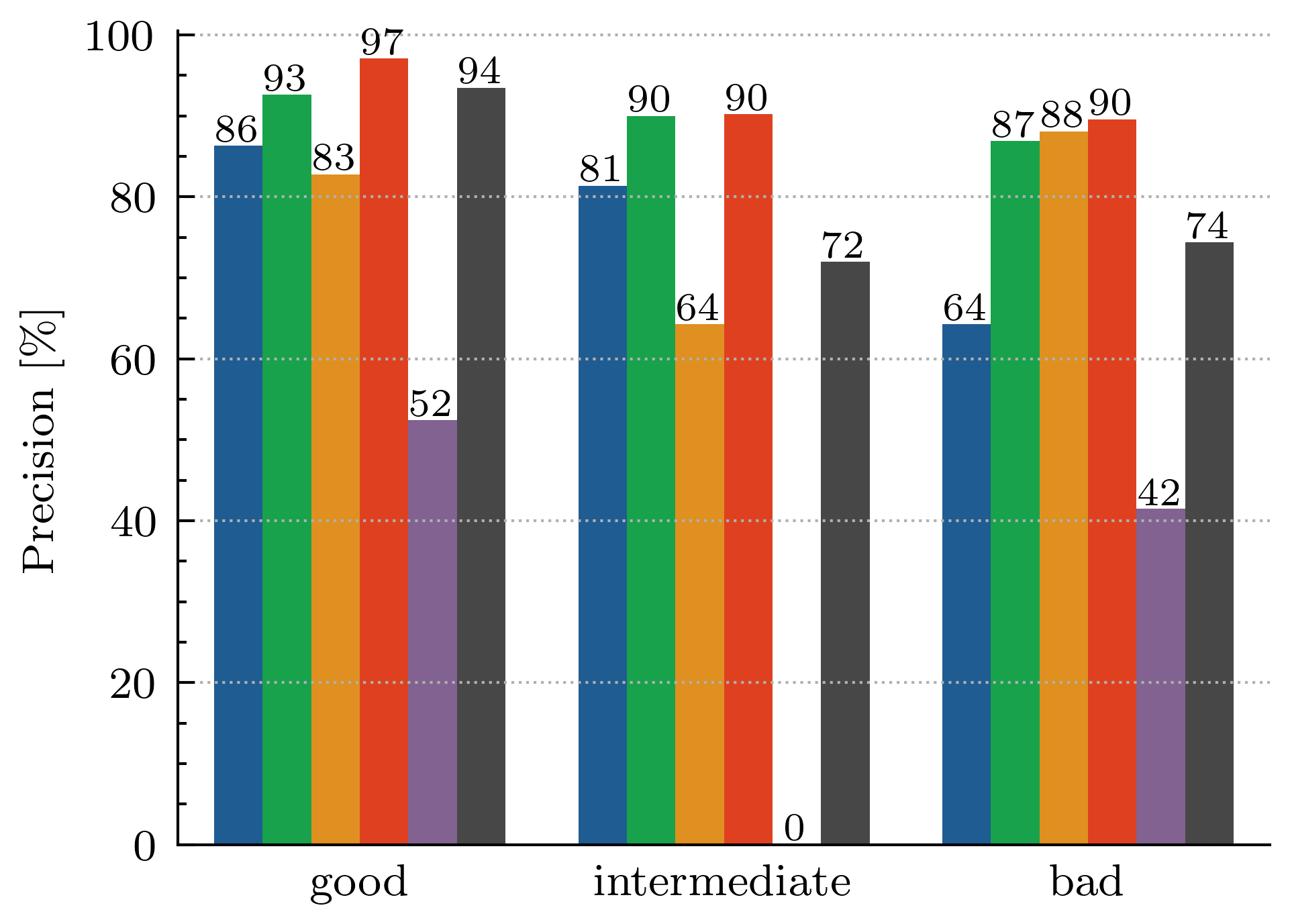}\label{fig:lr:prec}}
	\subfloat[Per class recall values for logistic regression.]{\includegraphics[width=0.4\linewidth]{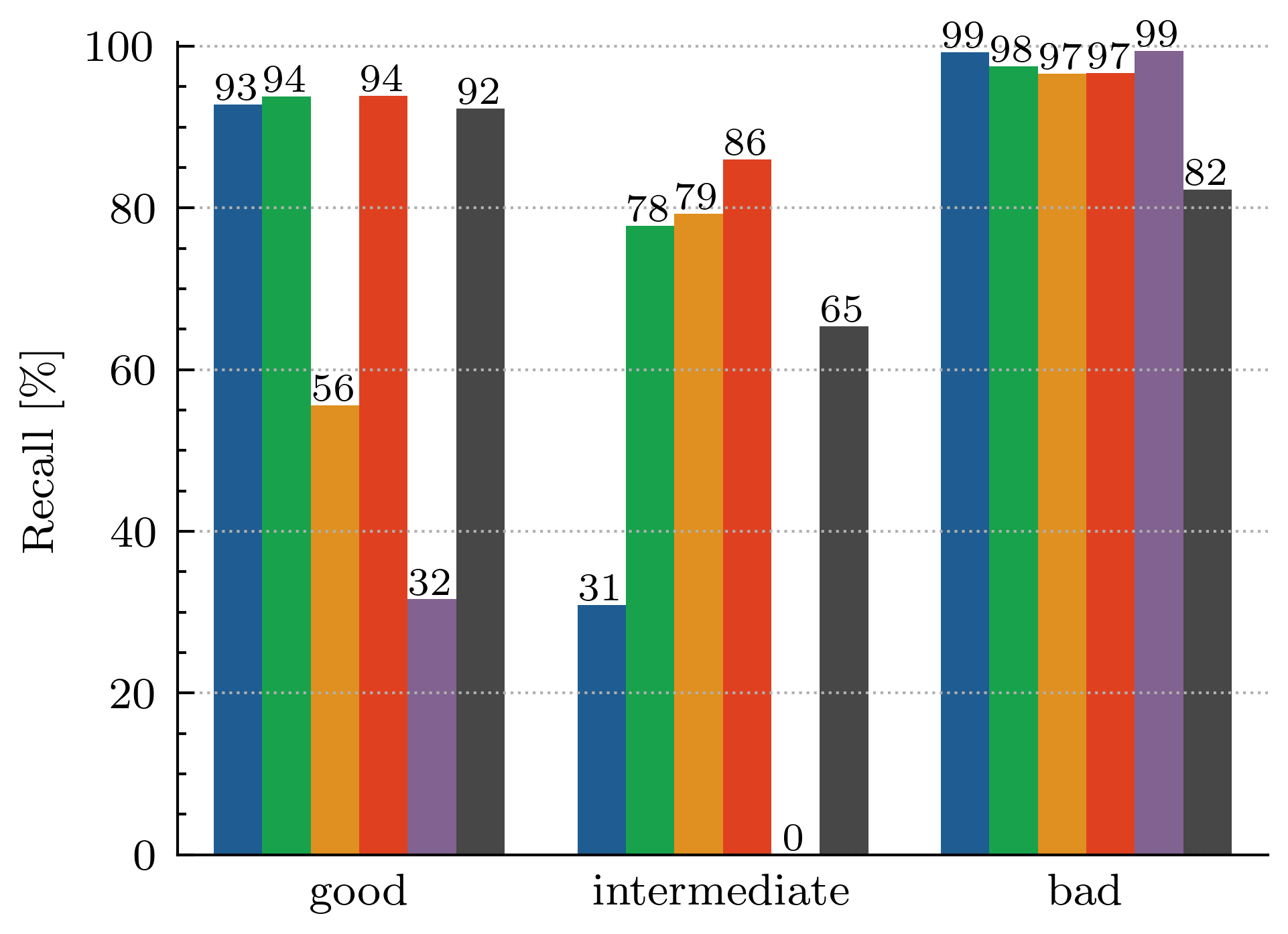}\label{fig:lr:rec}}%
	
	\subfloat[Per class precision values for decision trees.]{\includegraphics[width=0.4\linewidth]{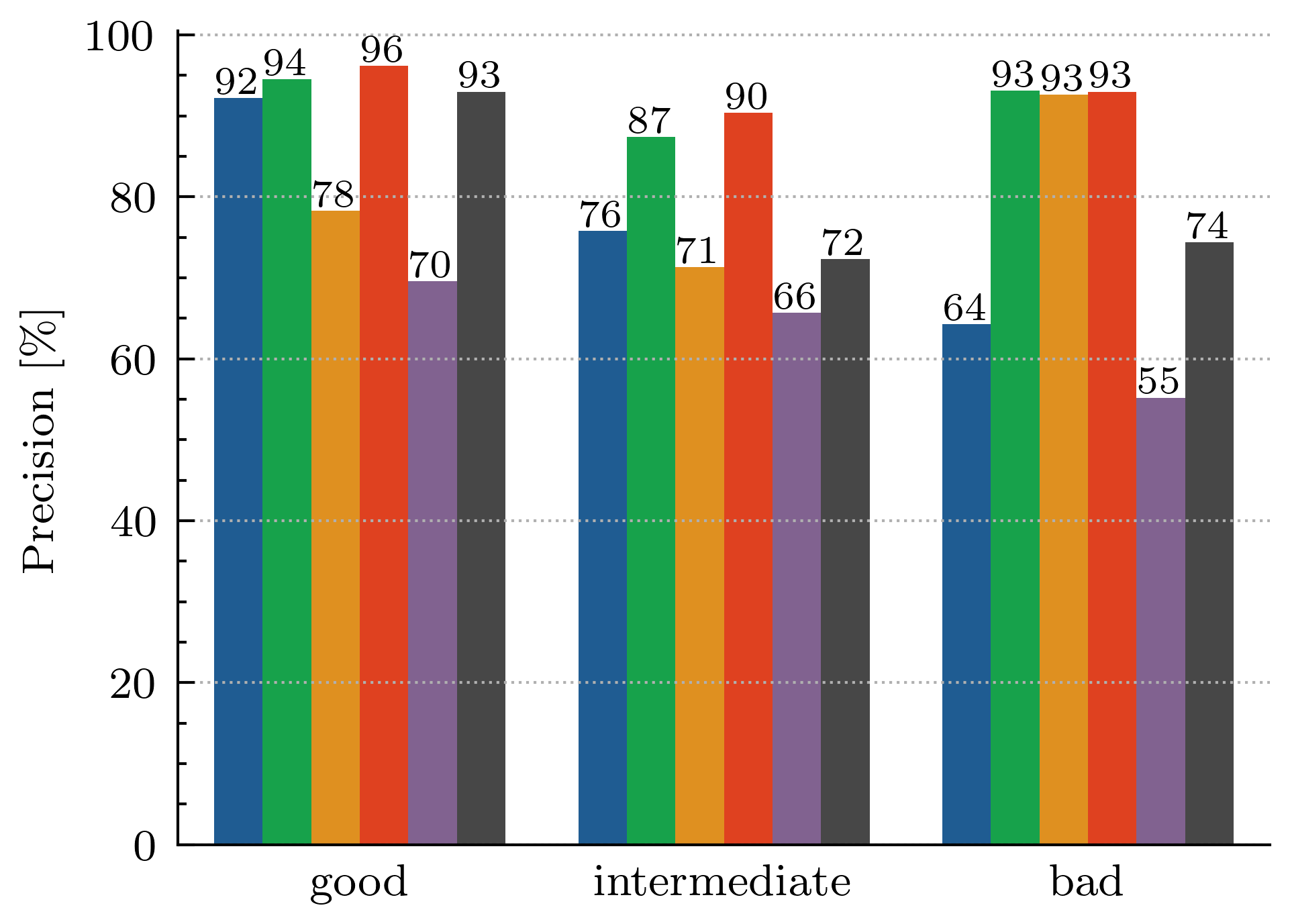}\label{fig:dt:prec}}
	\subfloat[Per class recall values for decision trees.]{\includegraphics[width=0.4\linewidth]{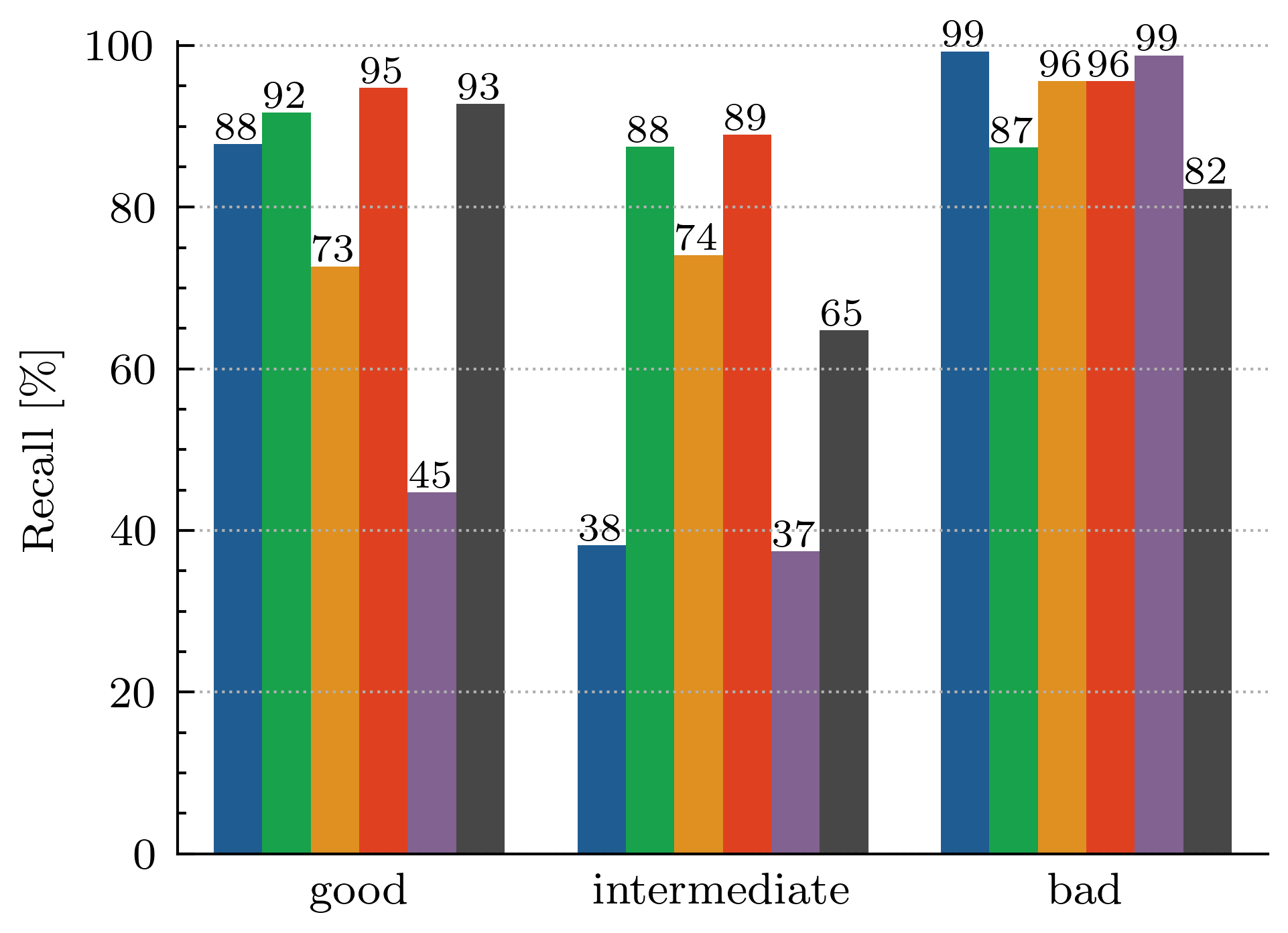}\label{fig:dt:rec}}%
	
	\subfloat{\includegraphics[width=0.8\linewidth]{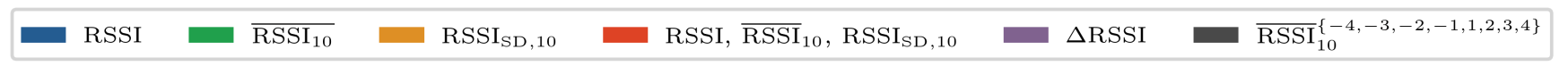}}
	\caption{Per class influence of the feature selection on fairness.}
	\label{fig:pc}
\end{figure*}

One of the widely-used approaches in ML for such trace-sets with small number of features is to examine whether synthetic features, such as average RSSI over a time window period or polynomial interactions~\cite{freitas2001understanding}, can assist the training to obtain more accurate models compared to that of the raw RSSI values. We study an extensive combination of features including 1) readily available RSSI, 2) averaged RSSI over 10 packets $\overline{\text{RSSI}_{10}}$, 3) standard deviation of RSSI over 10 packets RSSI$_{\text{SD},10}$, 4) a combination of the three RSSI, $\overline{\text{RSSI}}_{10}$, RSSI$_{\text{SD},10}$, derivate RSSI $\Delta$RSSI (``left'' derivative), and negative power of the averaged RSSI $\overline{\textrm{RSSI}}_{10}^{\{-4, -3, -2, -1, 1, 2, 3, 4\}}$ that are listed in Table~\ref{tab:features} and present the influence of the best-performing set of feature combinations on the classification performance. The table evaluates how well the learned model predicts link quality as per Eq.~(\ref{eq:prr-to-class}) for the next prediction window $\Wprr$, while relying on the parameters of Table~\ref{tab:parameters}.

The results show that only using $RSSI$ yields $74\%$ accuracy for the linear model and $75\%$ for the non-linear one as per the first line corresponding to each algorithm in Table~\ref{tab:features}. The F1 scores are $70\%$ and $72\%$ respectively, confirming the fact that accuracy overestimates the performance of the model on imbalanced datasets~\cite{jeni2013facing}. Breaking down into per class performance, it can be seen that F1 on the majority \textit{good} class is $70\%$ with a precision of $86\%$ and recall of only $93\%$ as also visually represented in Figures \ref{fig:lr:prec} and \ref{fig:lr:rec}. High precision and recall on this class show that the model is able to find the largest part of good links with minimal confusion. On the other hand, on the minority \textit{intermediate} class, the F1 is as low as ($44\%$) with a precision of $81\%$ and recall of only $31\%$. Low recall means that all the links detected as \textit{intermediate} are indeed intermediate, however only a small part of the total intermediate links in the data is detected. The model needs improvement to detect more \textit{intermediate} links accurately for better and fairer recognition of this minority class.

\begin{table*}[!th]
	\caption{Comparison of various data resampling strategies using linear and non-linear ML algorithms.}
	\label{tab:resample}
	\renewcommand{\arraystretch}{1.25}
	\centering
	\begin{tabular}{ l l l l l l }
		\toprule
		\bfseries Algorithm
		& \bfseries Resampling
		& \bfseries Acc. [\%]
		& \bfseries Precision [\%] 
		& \bfseries Recall [\%]
		& \bfseries F1 [\%]
		
		\\\midrule
		
		\multirow{3}{*}{Linear (Logistic)}
		& None
		& 96.8
		& 96.6 (98.8, 69.9, 98.9) 
		& 96.8 (99.0, 55.2, 98.6) 
		& 96.7 (98.9, 61.7, 97.4) 
		\\
		
		& RUS
		& 92.2
		& 92.3 (97.1, 90.2, 89.6)
		& 92.2 (93.9, 86.0, 96.7)
		& 92.2 (95.5, 88.0, 93.0)
		\\
		
		& ROS
		& 92.2
		& 92.3 (97.1, 90.2, 89.6)
		& 92.2 (93.9, 86.0, 96.7)
		& 92.2 (95.5, 88.0, 93.0)
		\\\midrule
		
		\multirow{3}{*}{Non-linear (DTree)}
		& None
		& 97.0
		& 97.0 (98.9, 66.9, 97.5)
		& 97.0 (98.6, 67.0, 98.1)
		& 97.0 (98.7, 67.0, 97.8)
		\\
		
		& RUS
		& 93.1
		& 93.1 (96.2, 90.2, 93.0)
		& 93.1 (94.6, 89.0, 89.6)
		& 93.1 (95.4, 89.6, 94.3)
		\\
		
		& ROS
		& 93.2
		& 93.2 (96.2, 90.4, 93.0)
		& 93.2 (94.8, 89.0, 95.6)
		& 93.2 (95.5, 89.7, 94.3)
		\\
		\bottomrule
	\end{tabular}
\end{table*}

Smoothing the $RSSI$ over 10 packets increases the performance to $89\%$ and $91\%$ respectively (line 2 in the table) while generating certain synthetic features further improves the results by 2-3 percentage points. Concretely, the fourth line corresponding to each algorithm in the table shows that learning from the feature set of {$RSSI$, $\overline{\text{RSSI}}_{10}$, RSSI$_{\text{SD},10}$} yields $92\%$ and $93\%$ accuracy, respectively. The high values of precision and recall for these feature combinations can also be visualized as in Figures~\ref{fig:lr:prec} and~\ref{fig:lr:rec}.

These results show that only using instant $RSSI$ as a feature with our imbalanced dataset is not sufficient to learn to discriminate the minority intermediate class sufficiently well. The F1 score for the \textit{intermediate} class is only $44\%$ for the linear model and $50\%$ for the non-linear model trained with $RSSI$ only. Similarly, also the precision and recall results for the intermediate class are modest for $RSSI$ only. As visualized in Figures \ref{fig:dt:prec} and \ref{fig:dt:rec}, precision is $76\%$ and recall is $38\%$ for the \textit{intermediate} class. 

When the two models are trained with a combination of features, namely {$RSSI$, $\overline{\text{RSSI}}_{10}$, RSSI$_{\text{SD},10}$}, the performance of the \textit{intermediate} class increases by more than $44\%$, resulting in a F1 score of $88\%$ for the linear model and $89\%$ for the non-linear model. This large increase in performance, leading to a fairer classification, also comes with slight increases of $1-2\%$ in the F1 scores of the majority classes. According to Figure \ref{fig:lr:prec} this feature combination results in a very good precision on all three classes for the linear model, namely $97\%$ on \textit{good} and $90\%$ on \textit{intermediate} and \textit{bad} respectively. For the non-linear number, the values depicted in Figure \ref{fig:dt:prec} are all very high as well, namely $96\%$ on \textit{good} and $90\%$ on \textit{intermediate} and $93\%$ on  \textit{bad} classes. It can be seen that the non-linear model is slightly more precise at determining \textit{bad} links with a slight penalty for \textit{good} links compared to the linear model. The recall values are also very high for both models. According to Figure \ref{fig:lr:rec}, the recall is $94\%$ on \textit{good} and $86\%$ on \textit{intermediate} and $97\%$ on \textit{bad} classes when the model is trained with the linear logistic regression, while Figure~\ref{fig:dt:rec} presents that the recall is $95\%$ on \textit{good} and $89\%$ on \textit{intermediate} and $96\%$ on \textit{bad} classes when the model is trained with the non-linear decision tree. It can be seen from these results that the advantage of the DTree model comes from its ability to yield higher recall values showing that not too many true positive have been missed in classification. While some of the \textit{intermediate} class links are still missed as there is about 10 percentage points difference compared to the other two classes (\textit{bad} and \textit{good}), {$RSSI$, $\overline{\text{RSSI}}_{10}$, RSSI$_{\text{SD},10}$} feature set provides the highest fairness.

The feature analysis also shows that by smoothing the training data, therefore removing noise and transitory fluctuations and capturing the boundaries of the variations, the learner can improve its performance and become fairer on the intermediate class. It is observed that the transient fluctuations are more prominent on the intermediate class, which is conforming to the findings of the literature~\cite{baccour2012radio}. 

\section{Compensating for the minority class in the training data to improve per class fairness}
\label{sub:resampling}

To compensate for the imbalanced class in the training data, and mitigate bias, the ML literature suggests employing resampling methods developed using statistical tools. These methods modify the distributions of the classes and re-balance the dataset. For our work, we consider two simple standard candidates; i) random oversampling (ROS), ii) random undersampling (RUS). The ROS~\cite{chawla2004special} approach considers duplicating the trace-set entries of the minority classes for all class sizes to reach the size of the majority class. The resultant resampled dataset is larger than the original. Contrarily, the RUS~\cite{chawla2004special} approach reduces all majority class sizes to the size of the minority class by randomly eliminating instances from other larger classes. Therefore, the obtained resampled dataset becomes smaller.

\begin{table*}
	\caption{The impact of linear and non-linear ML algorithms on the effectiveness of the ultimate LQE model.}
	\label{tab:algorithms}
	\centering
	\renewcommand{\arraystretch}{1.25}
	\begin{tabular}{@{}lllllll@{}}
		\toprule
		\bfseries Type
		& \bfseries Algorithm
		& \bfseries Acc. [\%]
		& \bfseries Precision [\%] 
		& \bfseries Recall [\%]
		& \bfseries F1 [\%]
		& \bfseries Training Time [s]
		
		\\\midrule

		Baseline
		& Majority classifier
		& 33.3
		& 11.1 (33.3, 0.0, 0.0)
		& 33.3 (0.0, 0.0, 0.0)
		& 16.7 (50.0, 0.0, 0.0)
		& 0.6
		\\
		
		\midrule
		\multirow{2}{*}{Linear}
		& Logistic regression
		& 92.2
		& 92.3 (97.1, 90.2, 89.6)
		& 92.2 (93.9, 86.0, 96.7)
		& 92.2 (95.5, 88.0, 93.0)
		& 2.5
		\\
		& SVM (linear kernel)
		& 92.1
		& 92.2 (97.4, 90.0, 89.2)
		& 92.1 (93.7, 85.8, 96.8)
		& 92.1 (95.5, 87.8, 92.8)
		& 93.6
		\\
		
		\midrule
		\multirow{2}{*}{Non-linear}
		& DTree
		& 93.1
		& 93.1 (96.2, 90.2, 93.0)
		& 93.1 (94.6, 89.0, 95.6)
		& 93.1 (95.4, 89.6, 94.3)
		& 1
		\\
		
		& MLP
		& 93.4
		& 93.4 (96.7, 90.5, 93.0)
		& 93.4 (94.9, 89.5, 90.0)
		& 93.4 (95.8, 90.0, 94.3)
		& 93.4
		\\
		\bottomrule
	\end{tabular}
\end{table*}
\begin{figure*}[htbp]
	\centering
	\subfloat[Multi-class ROC curve for logistic regression.]{\includegraphics[width=0.45\linewidth]{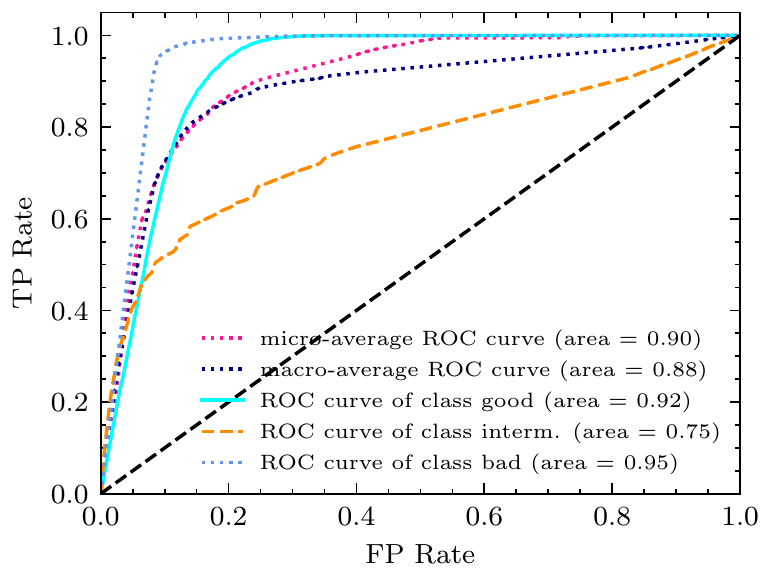}\label{fig:roc:logistic}}
	\subfloat[Multi-class ROC curve for decision trees regression.]{\includegraphics[width=0.45\linewidth]{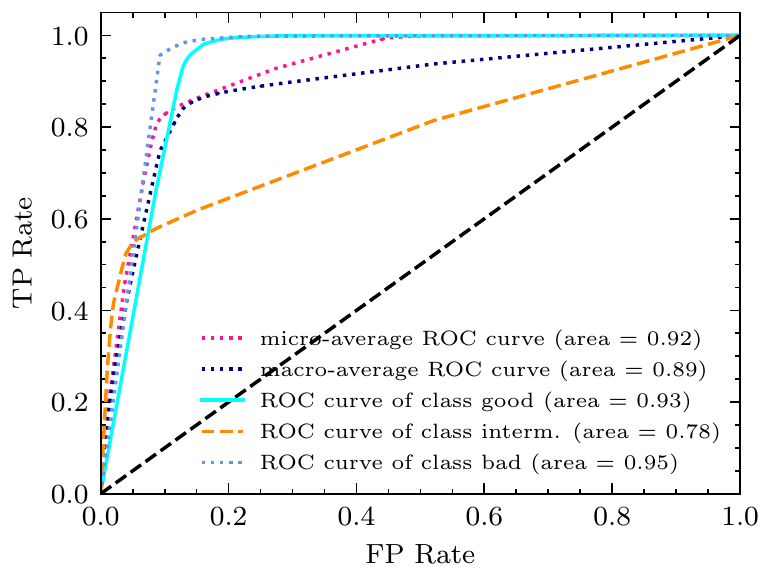}\label{fig:roc:dtree}}%
	\label{fig:roc}
	\caption{Multi-class receiver operating characteristic (ROC) representations portraying the performance of the two classification models.}
\end{figure*}

Table~\ref{tab:resample} presents the results of evaluation for the selected resampling strategies. For both classes of algorithms, it can be seen that employing RUS and ROS resampling strategies degrades the accuracy, precision, recall and F1 score. For the linear model and no resampling, the accuracy, precision, recall and F1 are $96\%$ and $97\%$ respectively. After employing resampling, they all drop by $4\%$ to $92\%$ and $93\%$ respectively. However, looking at the per-class break-downs in Table~\ref{tab:resample}, a more detailed insight can be acquired, where the performance discrimination on the majority classes decreases, expressively, the precision for the \textit{good} class drops from $98\%$ and $97\%$ for the linear model and from $98\%$ and $96\%$ for the non-linear model, while the precision for the \textit{bad} class drops from $98\%$ to $89\%$ for the linear model and from $97\%$ to $93\%$ for the non-linear model. However, the precision for the \textit{intermediate} class increases by over 30 percentage points from $69\%$ to $90\%$ for the linear model and from $66\%$ to $90\%$ for the non-linear model.  Similar conclusions can be drawn for the other metrics. 

The results in this section show that when optimizing the overall performance of the classifier, without consideration to per class fairness, the best results are obtained on the actual dataset resulting in $97\%$ accuracy, precision, recall and F1 scores. However, in this case the performance of recognizing the minority classes, namely \textit{intermediate} links is up to $67\%$ achieved by the non-linear model. In cases where correctly discriminating all classes is a requirement, then resampling is the recommended approach as it increases the correct discrimination, i.e., fairness, of the \textit{intermediate} links by over $20\%$.

\section{Performance evaluation of the model}
\label{sub:building-model}
We now compare the proposed DTree model with the other three ML models and a majority baseline, as summarized in Table~\ref{tab:algorithms}. In general, non-linear ML-based LQE models performed slightly better than the linear counterparts within a tiny margin of about $1\%$. This confirms the relatively non-linear nature of the problem and also verifies previous findings where linear regression (linear algorithm) and neural networks (non-linear algorithm) performed similarly~\cite{liu2014temporal}.

The tiny margin observed in Table~\ref{tab:algorithms} is also confirmed in Figs.~\ref{fig:roc:logistic} and~\ref{fig:roc:dtree}, where the figures present receiver operating characteristic (ROC) curve and the area under the curve (AUC) values for each of the class, and their micro and macro average performances. Indeed, non-linear model is slightly better due to a higher AUC value compared to that of the linear counterparts, for all link classes. This tiny margin is mainly due to the fact that in Rutgers trace-set, nodes are relatively close and in line-of-sight, and thus measurements data highly likely follow normal distribution. Contrarily, in case of non-line-of-sight and mobility scenarios, the input data would no longer follow any known statistical distribution. This is where non-linear counterparts, especially non-parametric algorithms, would be advantageous. For intermediate links, non-linear models outperformed the linear counterparts with about 2 percentage points margin. 

Considering computational complexity reflected in training time, as per the last column of Table~\ref{tab:algorithms}, we clearly demonstrate that the proposed LQE model based on DTree outperformed other LQE models in terms of computational complexity and at the same time, the DTree model accomplished one of the best performances for both the general model and the intermediate link class. DTree takes only 1 minute to train as opposed to 2.5 minutes for the logistic regression and it achieves slightly better performance ($1\%$). It takes 90 times less training time compared to MLP and the performance compromise is less than $1\%$.

\section{Summary and future work}
\label{sec:conclusion}
In this paper, we proposed a new decision tree based LQE model so as to improve fairness on minority classes. We compare the proposed classifier against three other ML approaches on a selected imbalanced dataset using five different performance metrics. Our study reveals that using additional metrics, such as F1 score to complement the widely used accuracy can help identify suboptimal performance on imbalanced datasets. For LQE, this means that the models are unfair and tend to confuse the \textit{intermediate}  quality links with \textit{bad} quality links. To this end, we demonstrated the impact of feature selection and resampling techniques on improving per-class classification. On the selected dataset, we showed that the performance on the minority class can be increased by over $40\%$ through feature selection and by over $20\%$ through resampling, leading to increased fairness. We also showed that non-linear models seem to be more appropriate for the problem, however, their advantage over linear models is marginal. Finally, we demonstrated that once training time is also taken into account, the proposed decision tree based model outperforms all the other considered models.

As a future work, we plan to extend the considered ML models to multi-technology LQE estimation as well as to use the recently developed LIME~\cite{lime} library for explainable deep learning to further investigate fairness aspects on such models.

\section*{Acknowledgment}
This work was funded in part by the Slovenian Research Agency (Grant no. P2-0016 and J2-9232) and in part by the EC H2020 NRG-5 Project (Grant no. 762013).

\bibliographystyle{IEEEtran}
\bibliography{ftr_eng}

\end{document}